\documentclass[oneside,ngerman]{article}
\usepackage[a4paper,top=3cm,bottom=3cm,left=4cm,right=4cm,marginparwidth=1.75cm]{geometry}
\usepackage{graphicx}
\usepackage{comment}
\usepackage{amsmath,amssymb}
\usepackage{color}
\usepackage{url}
\usepackage{hyperref}
\usepackage{xspace}
\usepackage{caption}
\usepackage{subcaption}
\usepackage{wrapfig}
\usepackage{multirow}
\usepackage{dsfont}
\usepackage{graphicx}
\usepackage{tikz}
\newif\ifreview
\reviewfalse

\ifreview
	\usepackage{lineno}

	\linenumbers
\fi

\title{\vspace{-2.5cm}{\bf Object detection characteristics in a learning factory environment using YOLOv8}}
\author{Toni Schneidereit, Stefan Gohrenz, Michael Breuß \\ 
\\[1ex]
\small Chair for Applied Mathematics (AI-Lab)\\ 
\small Brandenburg University of Technology Cottbus-Senftenberg \\
\small Platz der Deutschen Einheit 1, 03046 Cottbus, Germany \\
\small schneton@b-tu.de, StefanSebastian.Gohrenz@b-tu.de, breuss@b-tu.de
}

\date{\small \today} 

\begin{document}

\maketitle

\begin{abstract}
AI-based object detection, and efforts to explain and investigate their characteristics, is a topic of high interest. The impact of, e.g., complex background structures with similar appearances as the objects of interest, on the detection accuracy and, beforehand, the necessary dataset composition are topics of ongoing research. In this paper, we present a systematic investigation of background influences and different features of the object to be detected. The latter includes various materials and surfaces, partially transparent and with shiny reflections in the context of an Industry 4.0 learning factory. Different YOLOv8 models have been trained for each of the materials on different sized datasets, where the appearance was the only changing parameter. In the end, similar characteristics tend to show different behaviours and sometimes unexpected results. While some background components tend to be detected, others with the same features are not part of the detection. Additionally, some more precise conclusions can be drawn from the results. Therefore, we contribute a challenging dataset with detailed investigations on 92 trained YOLO models, addressing some issues on the detection accuracy and possible overfitting.
\end{abstract}
\begin{center}
\vspace{0.875cm}
\end{center}

\section{Introduction}
%
%
Artificial intelligence (AI) is of fundamental importance in Industry 4.0. The analysis of sensor data with AI can be utilised for the reliable recognition of complex patterns in real time, which is often a challenging task for humans \cite{zo:ch:sh:gu:ye}. For example, in predictive maintenance, AI may in this way help to identify and replace machine parts before they break.
More generally, main goals in predictive maintenance are to reduce production downtime and lowering the risk of damages in a factory \cite{ci:ab:ze:ko:as:sa,ca:so:vi:fr:ba:al,su:sc:pa:mc:be}, which may require an exact monitoring of the status of the factory and its processing of workpieces.
Other possible applications of AI in Industry 4.0 include robot automatisation, supply chain optimisation and quality control \cite{da:mc:re:ka,sa:cu:ka:kh:lo:po}. The latter is significant to maintain a high-level standard and to ensure that there are no harmful components or substances introduced into a production process. 

Companies are facing the challenge of adopting the concepts of Industry 4.0 in their operations. To foster this development, the use of learning factories may be considered.
A learning factory is a model in which learners can develop an understanding of practical problems from the real world, without tinkering with a real factory process \cite{ab:ch:si:me:el:se:si:el:hu:ti:se}.
In addition, to support teaching or training, such a factory may also be 
used to evaluate theoretical concepts from research before employing them in an industrial environment. The learning factories exist in different sizes and can be designed for different applications \cite{ab:ch:si:me:el:se:si:el:hu:ti:se,ab:me:ti:kr}.

%

\paragraph{\textbf{Related work}}
Especially in the context of predictive maintenance tasks and in optical quality control, vision-based methods and, more specifically, object detection techniques have been recognised as useful tools in manufacturing \cite{aj:eq:ma,wa:ch:qi:sn,li:su:ge:yi,ch:di:zh:zh:wu:sh}. Object detection is the computer vision task of localising and classifying objects in an image. AI-based detectors nowadays are oftentimes variants of convolutional neural networks and can be categorised into one-stage and two-stage detectors. Two-stage detectors, such as region-based Convolutional Neural Networks (R-CNNs) \cite{gi:do:da:ma} (including important variants like, e.g., Mask R-CNN \cite{he:gk:do:gi}), first extract region proposals and then classify them. This two-stage process (region proposal, classification) results in potentially very high accuracy at the expense of computational time. In contrast, one-stage detectors, such as YOLO (You Only Look Once) \cite{te:ce,re:di:gi:fa}, approach the object detection task as a regression problem. YOLO uses a single CNN-based framework that takes the entire image as input. Then YOLO simultaneously computes bounding boxes (localisation) and predicts class probabilities (classification) in a single pass. By doing this, YOLO models may achieve significantly higher computational efficiency compared to the two-stage process. For real-world applicability, one-stage detectors with the ability of real-time performance can be highly desirable. Therefore, the focus in this paper is on the YOLO model. 

The performance of a CNN is highly dependent on the availability of training data \cite{km:sch:my:br,ze:sch:fue:br}. The amount of images required can be reduced if the training set shows similar characteristics as the evaluation data. This was shown in an experiment in which the behaviour of sheep in a barn was observed \cite{ch:yu:wa:ca:li:zh}. The same camera angle was used for recording and evaluating the data, while also the background did not change. If it is not possible to collect sufficient data, datasets can be artificially expanded using augmentations \cite{bo:am,li:ta:li:li:ch}. This involves manipulating the original image, for example, by applying blur effects, noise or changes to the colour contrast, which can improve the robustness of a model \cite{sch:ya:sch:br:ge}.

The recognition of small objects has been identified as a major challenge for deep learning algorithms in several studies. One study compared the recognition accuracy of diseased rice plants with different YOLO models \cite{qu:qu:qu:ng}. The reason for the difficulties in recognizing small objects was found to be the small number of pixels, the image quality and a complicated background. As a result, only little information can be extracted.

\paragraph{\textbf{Problem statement and contribution}}
Our aim is to build upon and extend previous investigations on using the YOLO framework for object detection in a learning factory. More specifically, the authors of \cite{sch:ya:sch:br:ge} explore the suitability of several YOLO architectures distinguished in terms of size and complexity to monitor the process flow across a Fischertechnik Industry 4.0 application. An important issue studied in that work is the impact of different augmentation techniques in the construction of training datasets on detection quality. It turns out that a major point of interest is about the possible colour correlations between workpieces and factory background, influenced by interaction with lighting and light reflections as it is unavoidable in realistic scenarios. 

In this paper, we consider a selection of workpieces that feature highly challenging characteristics for object detection in an industrial environment. In addition to standard plastic workpieces in red, blue and white as employed in \cite{sch:ya:sch:br:ge}, we consider additional colours, as well as metallic, wooden and transparent materials. As in \cite{sch:ya:sch:br:ge}, the setting is a Fischertechnik Industry 4.0 learning factory, which represents an interesting setting and a controlled environment for studying computer vision applications like object detection \cite{ma:ri:se:kl:be}. 

Our contribution is a detailed study and analysis of the YOLO framework for use with datasets constructed at hand of this broad selection of workpieces in an Industry 4.0 learning factory environment. For this purpose, we have recorded a systematic and challenging dataset for each of the workpieces. There is only one parameter that differs between individual images of the datasets, which enables us to perform meaningful investigations. For the latter, a total of 92 YOLO models has been trained and evaluated. Possible overfitting and the explainability of different characteristics with, e.g., heatmaps, is part of the analysis. \par 
\medskip
The upcoming background section introduces the Fischertechnik learning factory with the workpieces, the YOLOv8 model with heatmaps, the performance metrics and the dataset creation process. This is followed by the experimental section, where results of the trained YOLOv8 single- and multi-class models are presented and discussed. Finally, in the summary and conclusion part, the findings are again summarised and discussed in a global context, together with an outlook on future work.

\section{Background and dataset}

\begin{figure}[!h]
    \centering
    \includegraphics[scale=0.75]{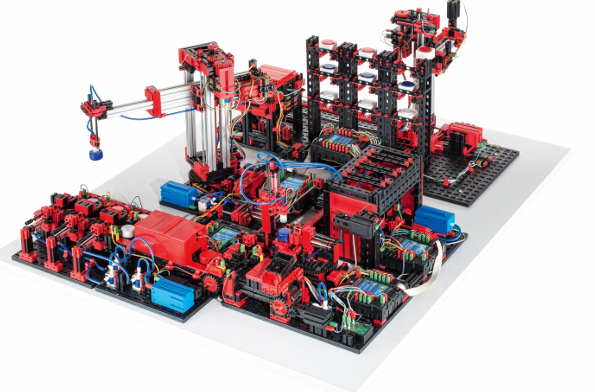}
    \caption{General overview of the Fischertechnik learning factory. Source: \cite{LFBilder}}
    \label{fig:FT_LF}
\end{figure}

\begin{wrapfigure}[9]{r}{0.6\textwidth}
\vspace*{-0.25cm}
\centering
\includegraphics[scale=0.25]{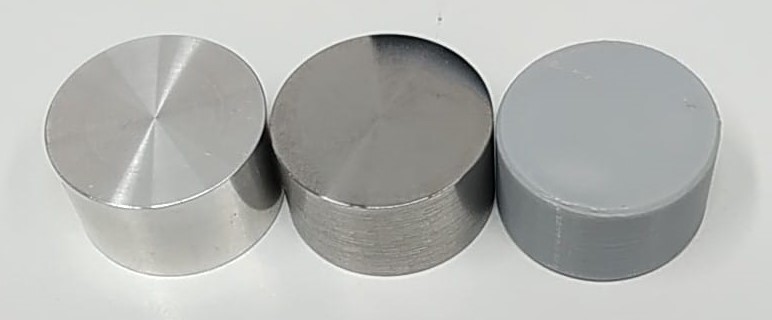}
\caption{Similarities in the appearance of the (left) aluminium, (mid) steel, (right) grey workpieces.}
\label{fig:SimilarGrey}
\end{wrapfigure}

The Fischertechnik learning factory simulates an Industry 4.0 fabrication cycle and is pictured in Fig.\ \ref{fig:FT_LF}. 
This model environment can be utilised for teaching and research purposes in various fields of, e.g., process automatisation and (AI-based) process monitoring. While no manufacturing takes place, the model simulates a production facility throughput. The factory consists of five modules, each one having its own controller-device which can be programmed in Python with the "ftrobopy" library \cite{ftrobopy}. The core module is a three-axis vacuum gripper, that transports so called workpieces (red, white, blue by default) through the factory. An oven and a milling station simulate processing tasks, while a conveyor belt with a colour sensor and sorting section categorise the workpieces by colour. Afterwards, the sorted workpieces can be stored in the high-storage warehouse. 
\begin{figure}[!h]
    \centering
    \includegraphics[scale=0.35]{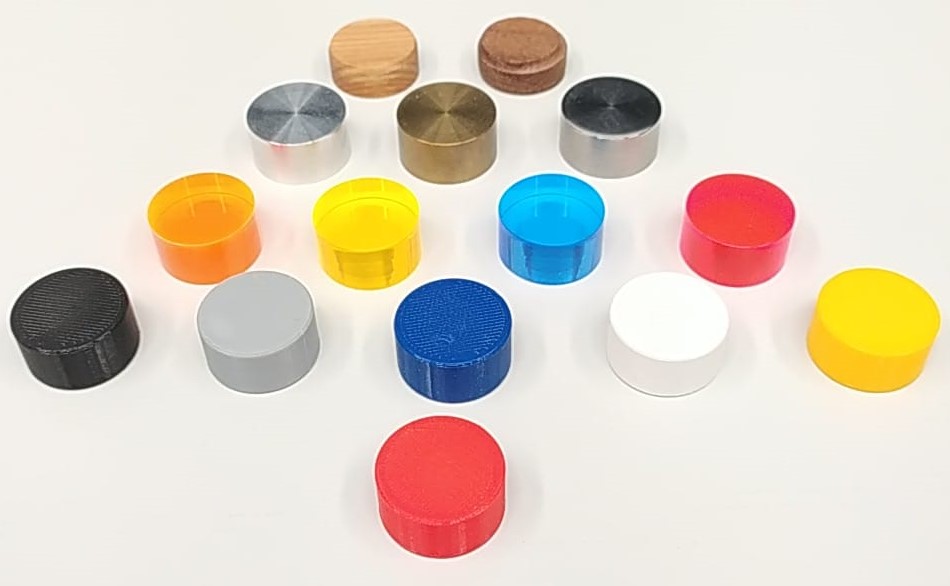}
    \caption{Overview of the 15 workpieces, from top to bottom: wood, metal, acrylic glass, PETG/plastic.}
    \label{fig:WP_additional}
\end{figure}
The cylindrical shaped workpieces come by default in the colours red, white and blue and are made of plastics. Along the proposed paper, we have created new workpieces of different colours and materials, which can be seen in Fig.\ \ref{fig:WP_additional}. These are made out of 3D printed PETG/plastic, fluorescent acrylic glass, different metals and wood. However, while the plastic, acrylic glass and metal workpieces are precisely dimensioned to match the default ones, the wooden pieces are in fact simple knothole dowels. The chosen metals are aluminium, brass and steel. Aluminium is a material that is used in the training factory for the suction pad supports. Depending on the incidence of light, the steel, aluminium and grey workpiece made of plastic may resemble each other, as pictured in Fig.\ \ref{fig:SimilarGrey}. 
In addition, the base plate of the learning factory is also grey. Brass was chosen as it has rough similarities to the yellow plastic and acrylic workpieces. The 3D printed black workpiece fits the many black parts of the factory.

\begin{figure}[!h]
    \centering
    \includegraphics[width=0.75\linewidth]{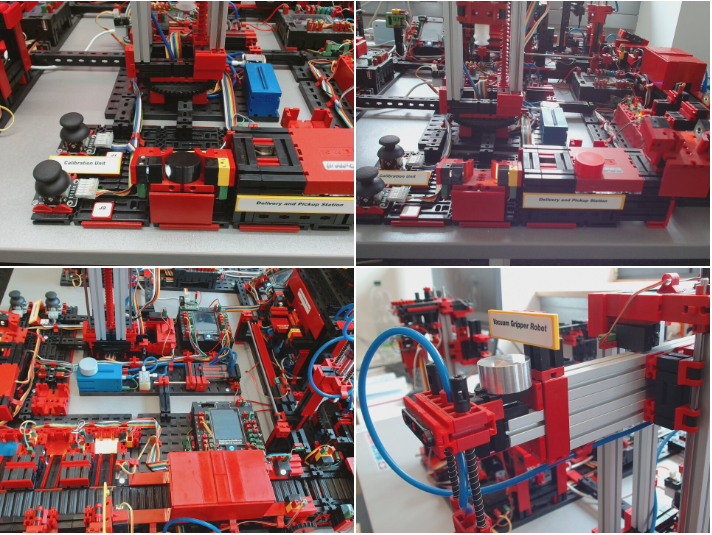}
    \caption{Examples of certain workpieces showing similar characteristics as components of the learning factory.}
    \label{fig:Similar}
\end{figure}

The described similarities motivated us to produce exactly these workpieces and to initiate a detailed investigation on the impact of colour and materials for object detection in an environment with similar characteristics. Some of the above-mentioned similarities can be seen in Fig.\ \ref{fig:Similar}.
\subsection{YOLOv8 and layer-wise relevance propagation}
The CNN-based framework of YOLOv8 (You Only Look Once version 8) is capable of performing real-time object detection and is therefore widely used \cite{te:co:ro}. In contrast to the so-called two-stage detectors, where the localisation and classification are performed in two separate steps, YOLO in general is referred to as a one-stage detector. The model views the task of object detection as a regression problem, while it processes an image through a single (but rather complex) CNN \cite{so:sa:ra}. 

In general, YOLO conceptually divides the input image into a grid, with each cell responsible for predicting objects within its area. A CNN architecture extracts crucial features from the image, enabling the model to identify and localise objects. While doing so, bounding boxes (center x, center y, width, height, confidence) together with a class and its probability are predicted for each cell. Finally, a non-maximum suppression is employed to filter multiple, redundant bounding boxes and to select the most confident predictions for each object. 

\begin{wrapfigure}[31]{r}{0.6\textwidth}
\vspace*{-0.25cm}
    \includegraphics[width=0.55\textwidth]{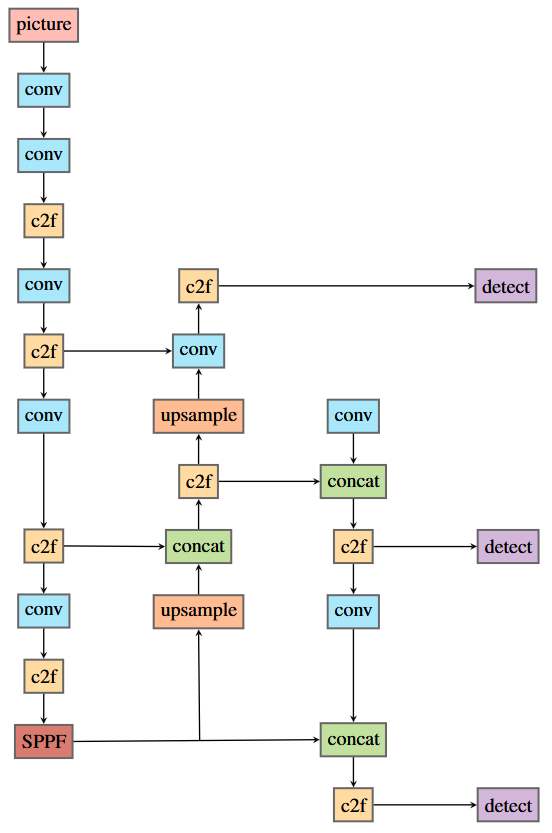}
  \caption{Schematic representation of the YOLOv8 architecture with different modules and multi-scale detection. Source: \cite{so:sa:ra}}
  \label{fig:YOLOarchitecture}
\end{wrapfigure}

The sketch of the corresponding network architecture is shown in Fig.\ \ref{fig:YOLOarchitecture}. A modified CSPDarknet53 architecture is used as the backbone. It consists of
53 convolutional layers that extract relevant information from the image. The c2f module aims to include context information of the image in the recognition process to potentially improve prediction accuracy \cite{so:sa:ra}. It is in general, according to the developers, based on a faster implementation of the CSP (Cross Stage Partial) module with two convolutions \cite{re:ku:ho:da}. The SPPF (Spatial Pyramid Pooling Fast) module is a more efficient implementation of the SPP module \cite{he:zh:re:su}, effectively capturing multi-scale features by dividing the feature map into a set of regions and applying pooling operations within each region. The generated features of the backbone are passed to the neck module, where they are merged. The result is passed on to the head module, which is responsible for localisation and the prediction of the bounding box. To prevent the loss of context information, the head module contains upsample layers that improve the resolution of the feature maps. YOLOv8 utilises an anchor-free approach, simplifying the model and potentially contributing to faster training and inference. YOLOv8 offers pre-trained models for detection, segmentation, and pose recognition \cite{ULYOLOv8,ULYOLOv11}. Different architectural sizes are available, providing different ratios of detection speed and accuracy.

\paragraph{\textbf{Layer-wise relevance propagation and YOLOv8 heatmaps}}
A trained CNN framework for object detection should be robust and reliable in its predictions. However, results are never one hundred percent correct. A visualisation of learned features and detections may improve both the explainability and the performance. One explanatory model is the LRP method (layer-wise relevance propagation) \cite{mo:bi:la:sa}. In this method, the output result is propagated backwards using local derivation rules. This distributes the information of a neuron evenly to the previous layer. The process is terminated when the input features are reached. LRP rules serve as parameters that manipulate the calculation of the derivation rules. This allows different layers of the network to be highlighted. Conclusions can be made from the results of these layers as to what the model recognises and which environmental factors influence the statement. 

In order to explain why some workpieces are recognized worse than others, the recognition process of the model is examined. The easy\_explain tool \cite{easy-explain} is used for this. This program creates a so-called heatmap. In this map, the areas that are frequently activated and react accordingly to learned features are marked in red. The algorithm is based on the layer-wise relevance propagation principle. This is intended to provide information to which features the model reacts on these test workpieces.

\paragraph{\textbf{Performance measurement metrics}}
In order to test the results of the later trained and evaluated YOLOv8 models, we use the mean Average Precision (mAP) metric in both variants mAP@[.5] and mAP@[.5:.95] \cite{pa:ne:si}. Both compare the ground truth bounding box and the predicted bounding box to compute the Intersection over Union (IoU). Depending on a given threshold for the IoU, a detection is either True Positive (TP), which means the IoU exceeded the threshold value, or False Positive (FP) when the predicted bounding box and the ground truth bounding box do not overlap enough. The false prediction of a correct instance is denoted by False Negative (FN). With those categories, we can define the metrics precision and recall. The former represents the ratio of correct predictions (TP) to all predictions (TP and FP). The latter is the ratio of correct predictions (TP) to all ground truth instances (TP and FN). A fixed IoU threshold and a fixed bounding box confidence threshold result in one pair of precision and recall values. In order to compute the so-called average precision (AP), which is defined as the area below the precision-recall curve, various pairs of precision and recall values are required. This is done by varying the bounding box confidence threshold and plotting the resulting precision over recall in a sorted way from 0 to 1. Since the IoU threshold also plays a role in the definition of TP, FP and FN, the AP is commonly specified into AP@[.5] (IoU threshold of 0.5) and AP@[.5:.95] (averaging the AP over different IoU thresholds, ranging from 0.5 to 0.95 in steps of 0.05). When an object detection model is capable of detecting more than one class, the different APs are averaged again into the mean average precision (mAP) metric. While mAP@[.5] more or less indicates whether the object was detected at all, mAP@[.5:.95] provides more information about how well the object was localised.

\subsection{Dataset creation and YOLO training}
In this section, we provide details about the data collection, plus the training and evaluation dataset composition, together with information about the applied image augmentation methods.

\begin{figure}[!h]
    \centering
    \includegraphics[width=1\linewidth]{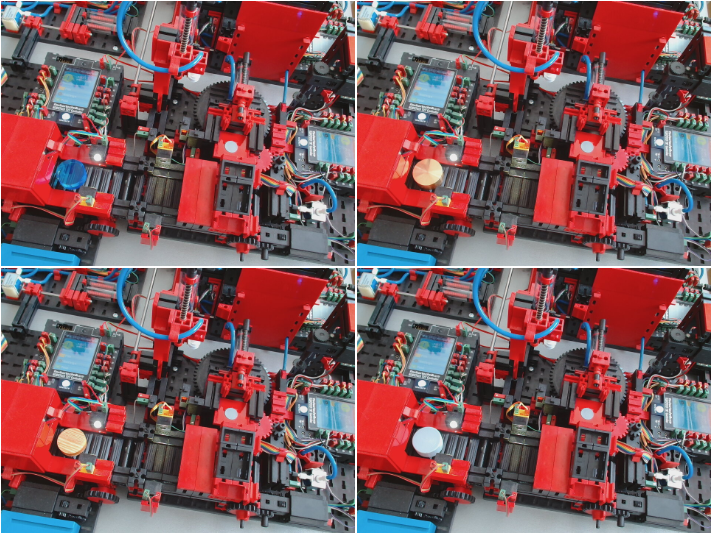}
    \caption{Example of one position with two different workpieces under the same point of view and lighting condition.}
    \label{fig:WP_pos1}
\end{figure}

\paragraph{\textbf{Data collection}}
The data base represents one of the core parts of our contributions. We have systematically created a dataset of 15 different workpieces for the Fischertechnik learning factory in order to investigate the impact of colour and appearance in the context of the environment. In addition to the choice of challenging materials and properties, the systematicity lies in the approach of recording the images. For each of the 5 learning factory models, each of the 15 workpieces has been recorded in 32 positions. While the camera point of view, the lighting condition and other environmental variables remained unchanged, all workpieces have been recorded in the same 33 positions. An example position of the blue acrylic glass and brass workpiece is given in Fig.\ \ref{fig:WP_pos1}.
Therefore, a total of 2,475 images have been recorded and labelled. The latter means manually drawing the bounding box around the object and saving the coordinates in the YOLO format (classID, centerX, centerY, width, height). We utilised the open source program "labelImg" \cite{Label}. Between recordings, the lighting conditions were achieved by switching the lights on and off or by raising and lowering the blinds. The camera positions were changed so that the workpieces were photographed from different angles and distances. When taking the pictures, we made sure that the learning factory was also visible as a background. These measures should lead to a high variance in the training images.

\paragraph{\textbf{Training and evaluation datasets}}
The entire dataset was separated into sub-datasets for individual trainings and investigations. A coarse division into individual training (120 images) and evaluation (45 images) datasets for every workpiece (15) is followed by a finer division of the training datasets into tiny (40 images), small (80 images) and medium (120 images). Every single image in the sub-datasets (both training and evaluation) shows the same positions of the individual workpiece, cf. Fig.\ \ref{fig:WP_pos1}. The tiny, small and medium training datasets are composed as follows: 
\begin{itemize}
    \item tiny: 40 individual positions
    \item small: tiny + 40 individual positions
    \item medium: small + 40 individual positions
\end{itemize}
We are totally aware that these numbers by no means are enough to create or train a fully robust object detection framework. The idea behind this approach is rather to find out, on which scale the single class detection accuracy changes, depending on the workpiece colour/material in the context of the learning factory background. On the other hand, we have also trained a YOLO model on the entire 2,475 images to compare the results.

However, in order to reduce possible false detection, we have augmented the datasets by different reasonable image processing techniques. That is, by copying an image and manipulating certain characteristics like, e.g., brightness, blur, translation, we artificially introduce aspects to the dataset that are perhaps not represented in the original images. Especially for the online object detection or the evaluation on previously unseen images or sequences, where different lighting conditions, camera angles and qualities are introduced, the augmentation may help to prepare the model, and enhance the robustness, for such situations. The python library "imgaug" \cite{imgaug1,imgaug2} was used to augment the images by the following 13 augmenters:
\begin{itemize}
    \item \textbf{contrast:} changes of the contrast value
    \item \textbf{BlendAlphaRegularGrid:} combines input image with grid-like cells with customizable alpha value
    \item \textbf{BlendAlphaElementwise:} creates greyscale of the image and blends result with original image with an alpha value
    \item \textbf{MotionBlur:} simulates motion blur
    \item \textbf{Solarize:} inversion of all pixel values below a threshold value
    \item \textbf{Rotation:} rotates image around image centre
    \item \textbf{shearX/shearY:} shears the image in X or Y direction 
    \item \textbf{AllChannelsCLAHE:} extracts intensity-dependent colour channel and applies contrast-related histogram to all colour channels
    \item \textbf{Canny:} blends detected edges with original image
    \item \textbf{Flipud/Fliplr:} flips the image vertically/horizontally
    \item \textbf{ElasticTransformation:} moves pixels within a defined area randomly
    \item \textbf{Affine:} scales the image
    \item \textbf{GaussianBlur:} blurs by applying the Gaussian function
\end{itemize}
All training datasets have been augmented in the same way, using the same parameters. Creating 120 augmented copies of each original image, with varying parameters, results in the final dataset sizes given in Tab.\ \ref{tab:dataset}. Therefore, e.g., the tiny training dataset consists of 40 original images plus 13 times 120 augmented images, resulting in a total amount of 1,600 training images. The small and medium datasets are composed in the same way. 
\begin{table}[!h]
\centering
\caption{Number of images in the training datasets for each of the 15 workpieces, before and after augmentation.}
\begin{tabular}{|c|c|c|}\hline
\textbf{dataset}& \textbf{original number of images} & \textbf{total number of images} \\\hline
tiny       & 40   & 1,600 \\
small      & 80   & 3,200 \\
medium     & 120  & 4,800 \\
multi-class & 1,800 & 72,000 \\
\hline
\end{tabular}
\label{tab:dataset}
\end{table}
We have trained YOLOv8 for each workpiece ("single class") and each dataset separately with YOLOv8n (nano) and YOLOv8x (extra large) for 100 epochs. As well as for the multi-class dataset, which includes all training images. In theory, depending on the architectural sizes, the nano version should be the fastest in terms of detection speed, while the extra large architecture should win the accuracy race. However, the larger the architecture, the more parameters have to be trained. In the upcoming section, we will investigate important aspects of the training, evaluation and prediction results.

\section{Experiments and Results}

The results in this section refer to the training and evaluation of YOLOv8 trained separately for each workpiece (single-class) and trained on all images (multi-class). Let us recall that for the experiments on the single-class detection, we trained each class on three different dataset sizes (tiny - 1,600 images, small - 3,200 images, medium - 4,800 images) and evaluated everything on 45 unseen images. All the parameters regarding the composition of the training and evaluation datasets in terms of position in the learning factory, camera angle and lighting condition per image are equal per workpiece. Therefore, the only parameter that changes per image is about the appearance of each workpiece. Everything related to the multi-class detection was trained on 72,000 images and evaluated on 675 unseen images.

\subsection{Single-class training and evaluation}

\begin{figure}[!h]
    \centering
    \includegraphics[scale=0.35]{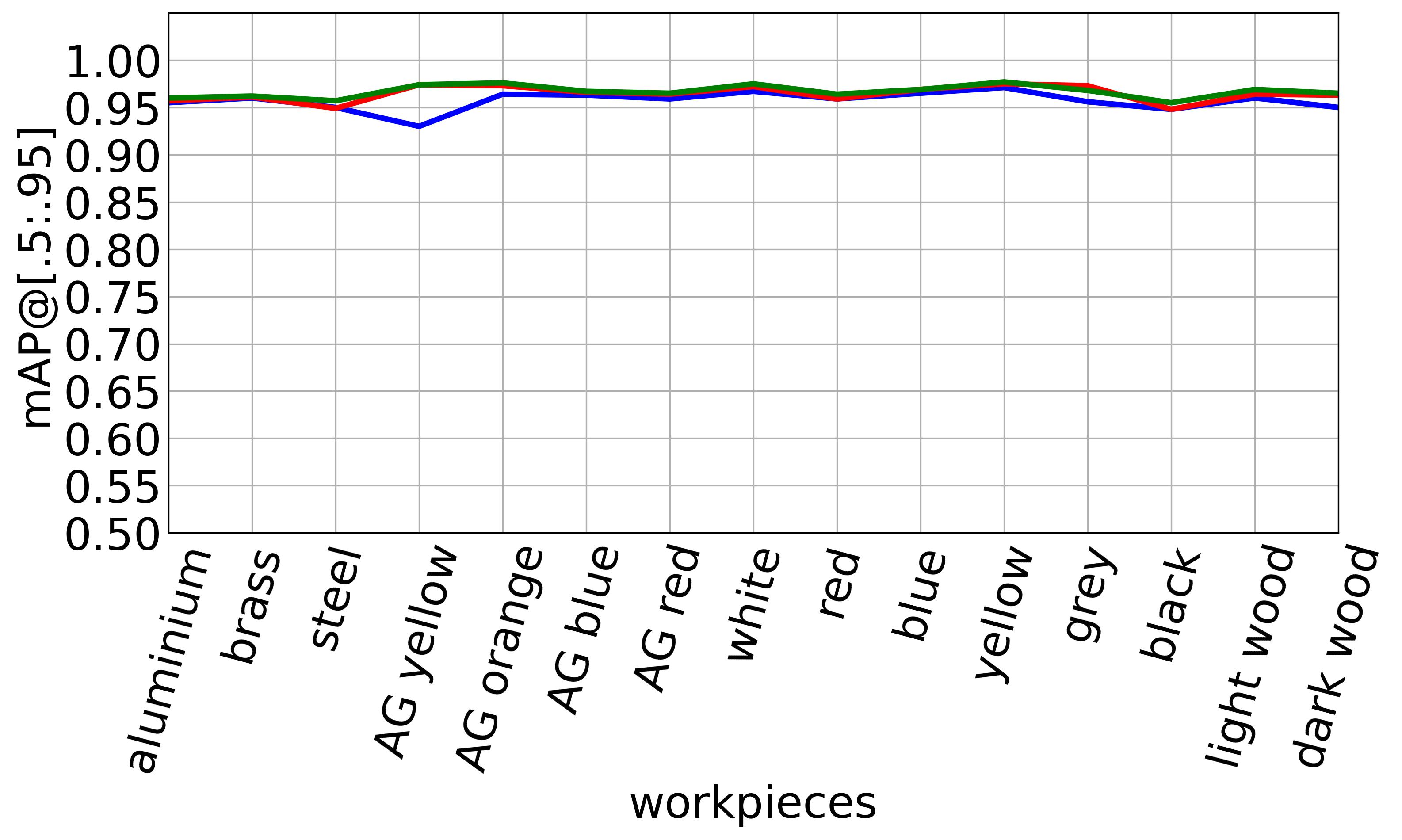}
    \caption{Training mAP for the 15 separately trained YOLOv8n models, where AG is an abbreviation for acrylic glass, (blue) tiny dataset, (red) small dataset, (green) medium dataset.}
    \label{fig:SC_YOLOv8n_Train}
\end{figure}
\paragraph{\textbf{Training results}}
The training mAP@[.5:.95] for the 15 single-class YOLOv8n models in Fig.\ \ref{fig:SC_YOLOv8n_Train} already shows quite good results, where only the yellow acrylic glass workpiece is an outlier in the tiny dataset. Apart from that, all three dataset sizes show a similar behaviour.
\begin{figure}[!h]
    \centering
    \includegraphics[scale=0.35]{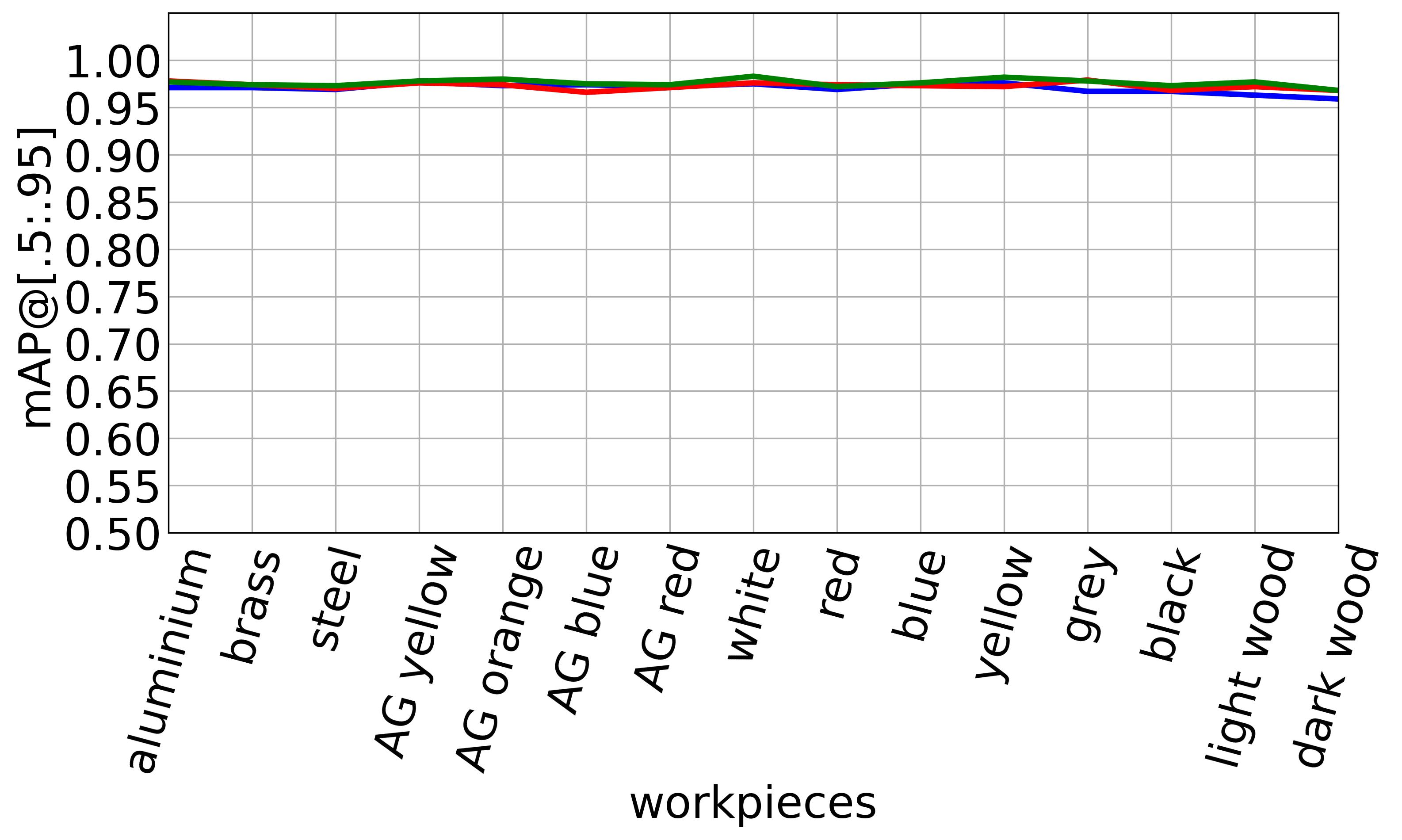}
    \caption{Training mAP for the 15 separately trained YOLOv8x models, where AG is an abbreviation for acrylic glass, (blue) tiny dataset, (red) small dataset, (green) medium dataset.}
    \label{fig:SC_YOLOv8x_Train}
\end{figure}
This trend also holds for the training mAP@[.5:.95] of YOLOv8x in Fig.\ \ref{fig:SC_YOLOv8x_Train}. Here we find even more accurate results, as one would expect from the architectural size of the extra large model. There is accuracy-wise almost no difference between the tiny, small and medium datasets.  
\begin{figure}[!h]
    \centering
    \includegraphics[scale=0.35]{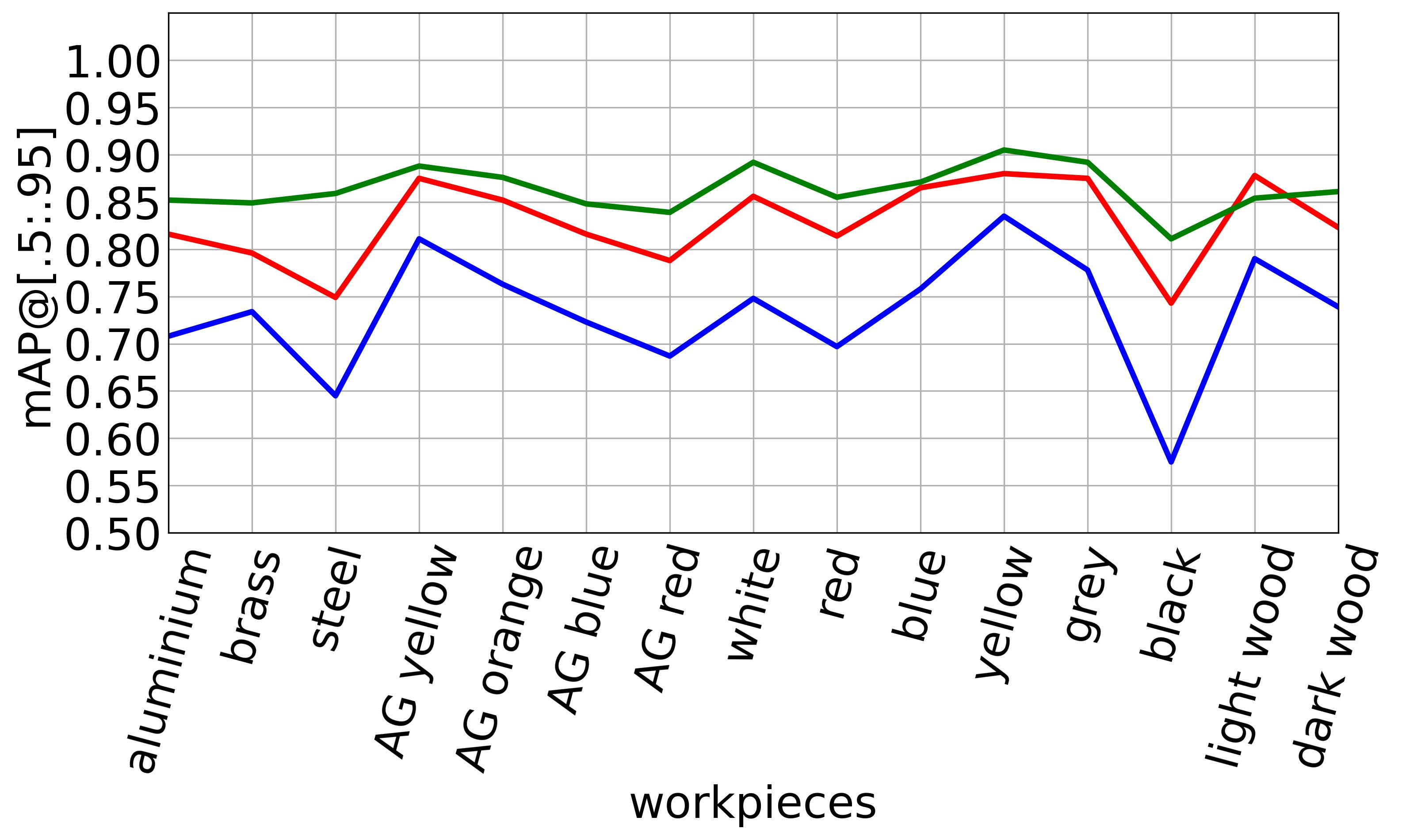}
    \caption{Evaluation mAP for the 15 separately trained YOLOv8n models, where AG is an abbreviation for acrylic glass, (blue) tiny dataset, (red) small dataset, (green) medium dataset.}
    \label{fig:SC_YOLOv8n_Eval}
\end{figure}

\begin{figure}[!h]
    \centering
    \includegraphics[scale=0.35]{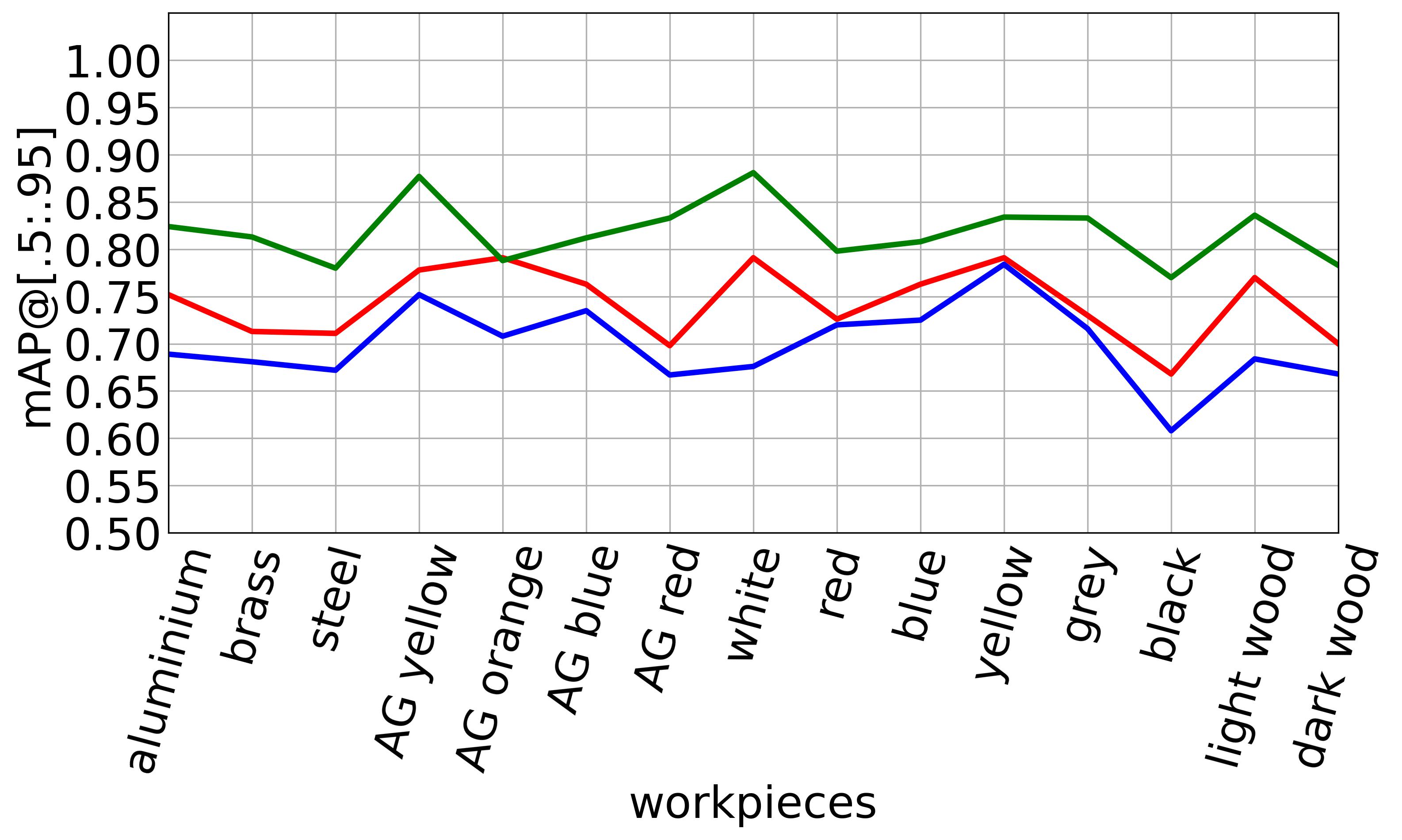}
    \caption{Evaluation mAP for the 15 separately trained YOLOv8x models, where AG is an abbreviation for acrylic glass, (blue) tiny dataset, (red) small dataset, (green) medium dataset.}
    \label{fig:SC_YOLOv8x_Eval}
\end{figure}
\paragraph{\textbf{Evaluation results}}
However, more meaningful and interesting are the testings of the single-class trained YOLOv8 models on unseen evaluation images (45 per workpiece). Those results are shown in Fig.\ \ref{fig:SC_YOLOv8n_Eval} for YOLOv8n and in Fig.\ \ref{fig:SC_YOLOv8x_Eval} for YOLOv8x. In contrast to the training results, differences between the three datasets are partially significant, although the general trend for YOLOv8n is very similar with respect to the dataset sizes. We find the bounding box prediction to be most accurate for the workpieces that do not share the same colours as the factory background. Those are both the yellow (plastic and acrylic glass) coloured workpieces. On the other side, the black plastic piece shows the worst mAP@[.5:.95], most likely because of the many black factory parts. This statement partially holds for both red classes (plastic and acrylic glass). However, although the baseplate is of grey colour, the corresponding plastic workpiece shows a better performance, compared to black and red. Therefore, the relation between the class and background colour can not be the only important component of the detection performance. This general trend continues for the larger datasets, with better mAP@[.5:.95] values.

Turning to the evaluation for YOLOv8x in Fig.\ \ref{fig:SC_YOLOv8x_Eval}, the differences between the datasets are similar. The larger the dataset, the better the mAP@[.5:.95] values. However, in contrast to YOLOv8n, we find some significant differences in the general trend. The bounding box prediction of the red plastic class has significantly improved for the tiny dataset (blue). This is particularly interesting, since the results for the red acrylic glass workpiece did not improve. This, however, does not hold for the larger datasets. So the performance regarding the red colour has partially improved, which implies, that the extra large architecture seems to focus more on additional characteristics besides the colour. Black still has the worst performance. A highly interesting aspect is the fact, that the general trend of the medium dataset is less smooth, compared to Fig.\ \ref{fig:SC_YOLOv8n_Eval}.
\begin{figure}[!h]
    \centering
    \includegraphics[scale=0.35]{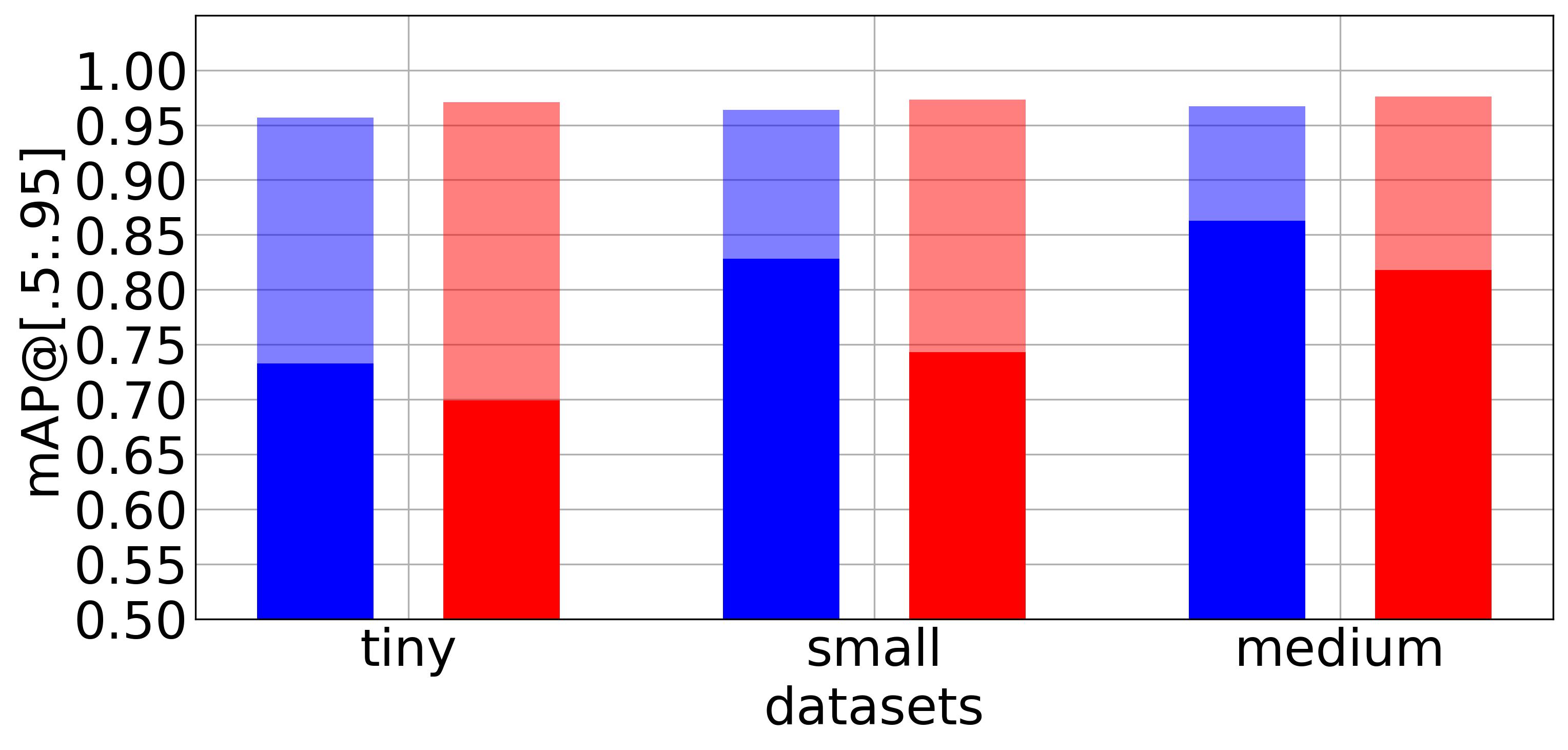}
    \caption{Training (transparent) and evaluation (solid) mAP averaged over all 15 workpieces, (blue) YOLOv8n, (red) YOLOv8x.}
    \label{fig:SC_bar_EvalTrain}
\end{figure}
\paragraph{\textbf{Averaged results for training an evaluation}}
In Fig.\ \ref{fig:SC_bar_EvalTrain} we have displayed the mAP@[.5:.95] values, averaged over all classes, for both training (transparent) and evaluation (solid) for YOLOv8n (blue) and YOLOv8x (red). Although the training performance of YOLOv8x is slightly better than YOLOv8n, the evaluation performance is much worse. The difference is most significant for the small dataset, while the medium dataset seems to catch up again. Nonetheless, in this experiment, the YOLOv8n model clearly outperforms the largest architecture. This behaviour can be an indication of overfitting, although the training and validation loss from the training process (not shown here) do not confirm this assumption. However, another possible explanation can be an insufficient size or composition of the datasets, although 4,800 training images (as well as the training mAP@[.5:.95]) should be arguably enough. 
\paragraph{\textbf{Bounding box confidence}}
In Tab.\ \ref{tab:BBconf_80} we listed results related to the bounding box confidence, which is computed by the YOLO framework. This confidence value indicates how certain the model is that an object is located within the detected bounding box. As the models were each trained on a single-class, the results can be clearly assigned to an object. The results given in Tab.\ \ref{tab:BBconf_80} now specifically show the number of predictions below a confidence of 80\% for some of the worst performing workpieces. Besides a confirmation for the black workpiece to perform poorly and both red classes being among the worst results, the table shows an interesting contrast to the previous findings. While in Fig.\ \ref{fig:SC_bar_EvalTrain} YOLOv8n had a significantly better evaluation mAP@[.5:.95] than YOLOv8x, now we find that the transition from the smaller to the larger model reduces the predictions with a bounding box confidence below 80\%. That means the classification accuracy itself gets better with the larger model, while the localisation accuracy decreases.
\begin{table}[!h]
    \centering
    \caption{Absolute numbers of detections on the evaluation images (45 per class) with a bounding box confidence below 80\%, where AG is an abbreviation for acrylic glass. The results are from YOLOv8n with the medium dataset.}
    \begin{tabular}{|c|c|c|}\hline
        workpiece & YOLOv8n & YOLOv8x\\ \hline
         aluminium& 6 & 1\\
         steel& 8   & 2\\
         black& 10& 7 \\
         red& 6     & 3\\
         AG red& 7 &  1 \\
         AG blue& 4 & 2 \\
         \hline
    \end{tabular}
    \label{tab:BBconf_80}
\end{table}
\begin{figure}[!h]
    \centering
    \includegraphics[scale=0.3]{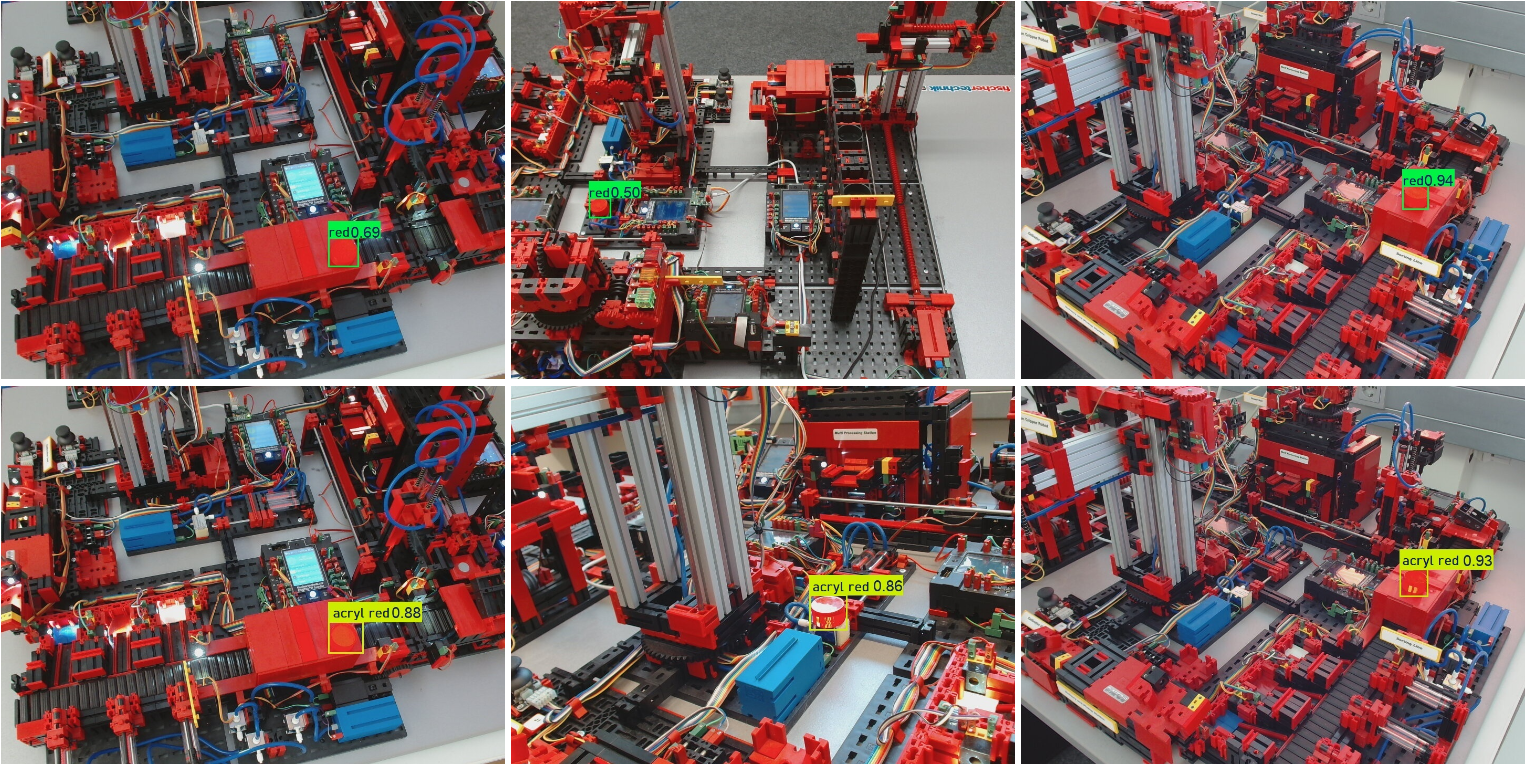}
    \caption{Prediction examples for the red (top row) and acrylic glass red (bottom row) workpieces, evaluated with YOLOv8n, trained on the medium dataset. Because of the challenging visibility, the confidence values are (top left) 0.69, (top mid) 0.50, (top right) 0.94, (bottom left) 0.88, (bottom mid) 0.86, (bottom right) 0.93.}
    \label{fig:PredResults}
\end{figure}
\paragraph{\textbf{Discussion of workpiece appearance}}
Now we want to discuss prediction results in the context of the workpiece appearance, which can be seen exemplarily in Fig.\ \ref{fig:PredResults}. The images show the red workpiece (top row) and the acrylic red workpiece (bottom row) with the prediction results from YOLOv8n. Although the left- and right-most images each show the same position and perspective, the results are non-consistent. Both workpieces in the left-most images show quite a difference in the prediction, although for the human eye, the appearance of the red and the acrylic glass red workpiece is very similar. At least compared to the right-most images, where both confidence values are almost the same, simply the camera angle has changed. Even the reflection at the acrylic glass workpiece seems to rather increase the score. Speaking of the reflection and considering the bottom mid image, the acrylic glass red class takes both the reflecting and non-reflecting characteristics into account. The top mid image of the red class, however, has a very poor confidence score, even worse than in the top left image, although the direct background has a greater variation of colours and shapes.  
\begin{figure}[!h]
    \centering
    \includegraphics[scale=0.4]{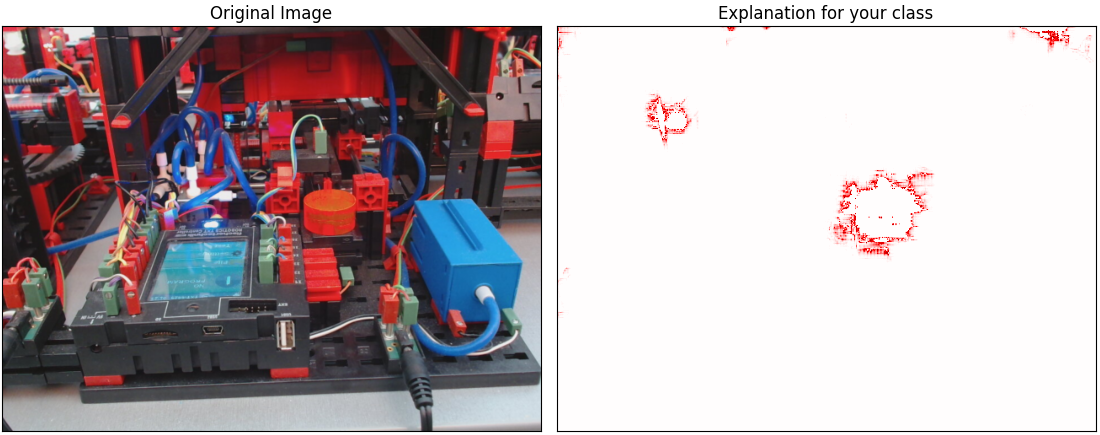} \\
    \includegraphics[scale=0.4]{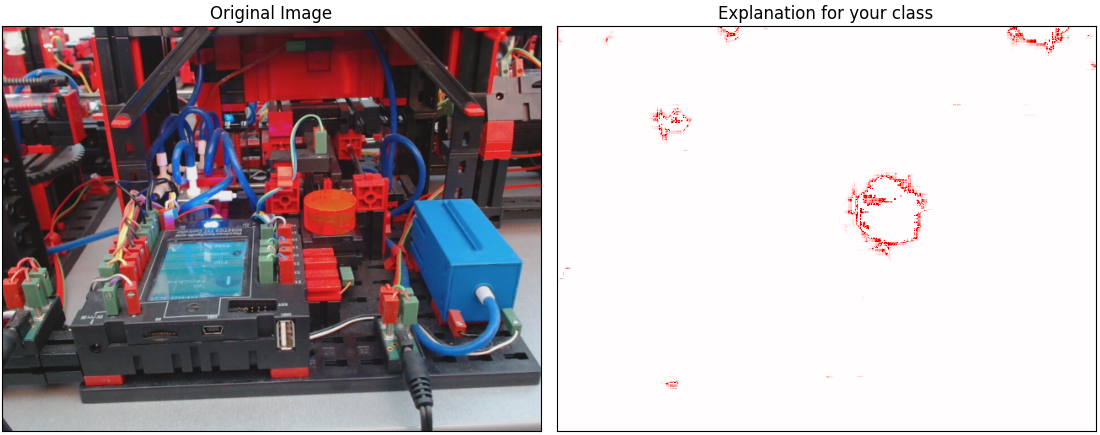}
    \caption{Heatmaps of an evaluation image for the red acrylic glass workpiece, trained on the medium dataset, (top) YOLOv8n, (bottom) YOLOv8x.}
    \label{fig:heatmap}
\end{figure}
\paragraph{\textbf{YOLOv8 heatmaps}}
In Fig.\ \ref{fig:heatmap}, two YOLOv8 heatmaps of the red acrylic glass workpiece are given. It achieved a confidence value of around 96\% in the evaluation image shown. A red connection, together with parts of the red wall section, was marked to the left of the workpiece. It is not possible to determine why the connecting element to the right of the workpiece was not part of the heatmap. The reason for the marking in the top right-hand corner cannot be clearly explained either. The model reacts to the red elements of the learning factory and therefore background elements seem to be relevant. This may indicate or verify that the YOLO framework reacts to more than the specific combination of shape and colour. While the heatmap of YOLOv8x seems to sharpen some of the element contours, also some new markings are introduced. Therefore, further investigations on the explainability, e.g., with heat- and feature maps, are necessary. 

\subsection{Multi-class training and evaluation}
\begin{figure}[!h]
    \centering
    \includegraphics[scale=0.45]{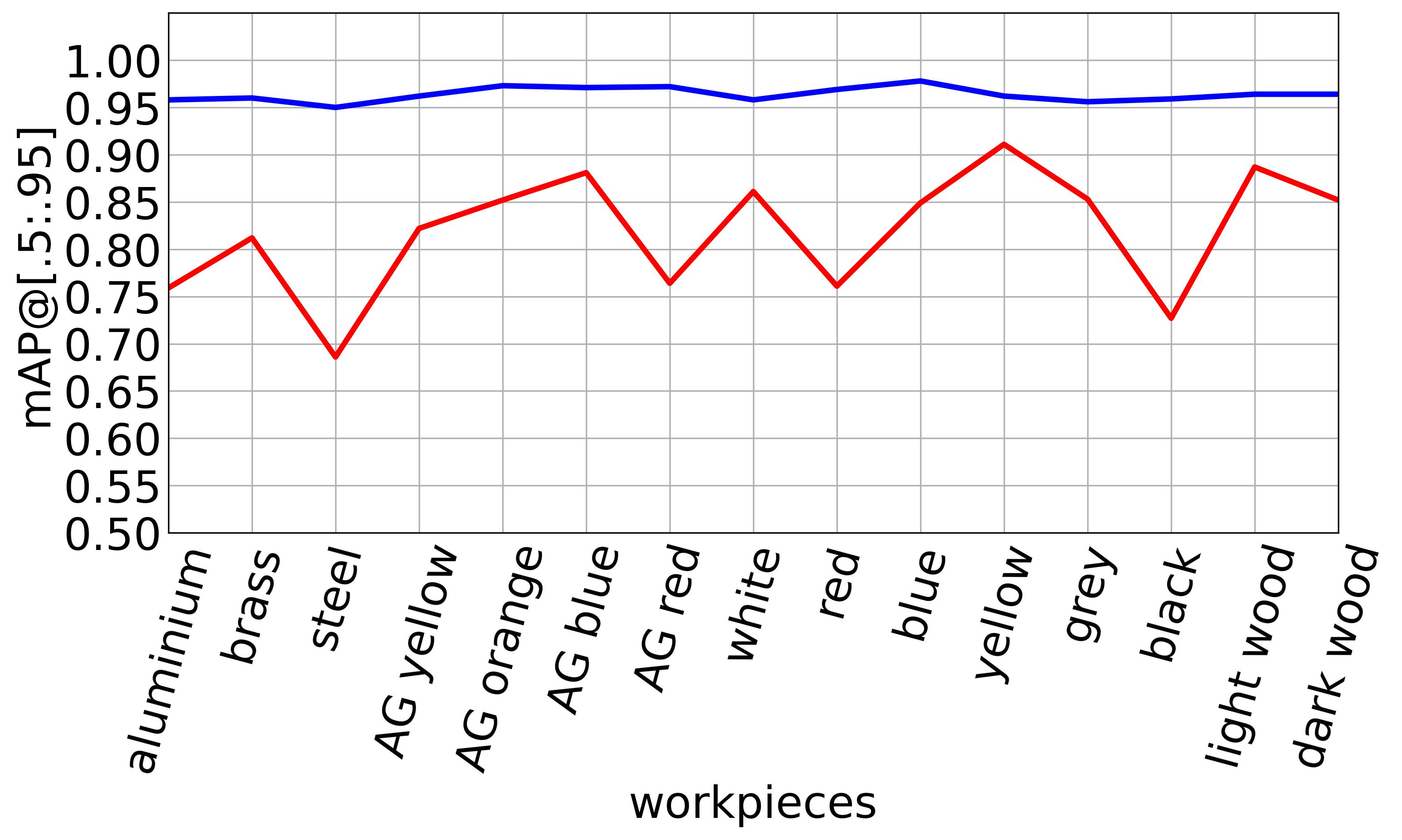}
    \caption{Results for the YOLOv8n multi-class model, trained on all the images at once, (blue) training mAP, (red) evaluation mAP)}
    \label{fig:MulticlassYOLOv8n}
\end{figure}
\begin{figure}[!h]
    \centering
    \includegraphics[scale=0.45]{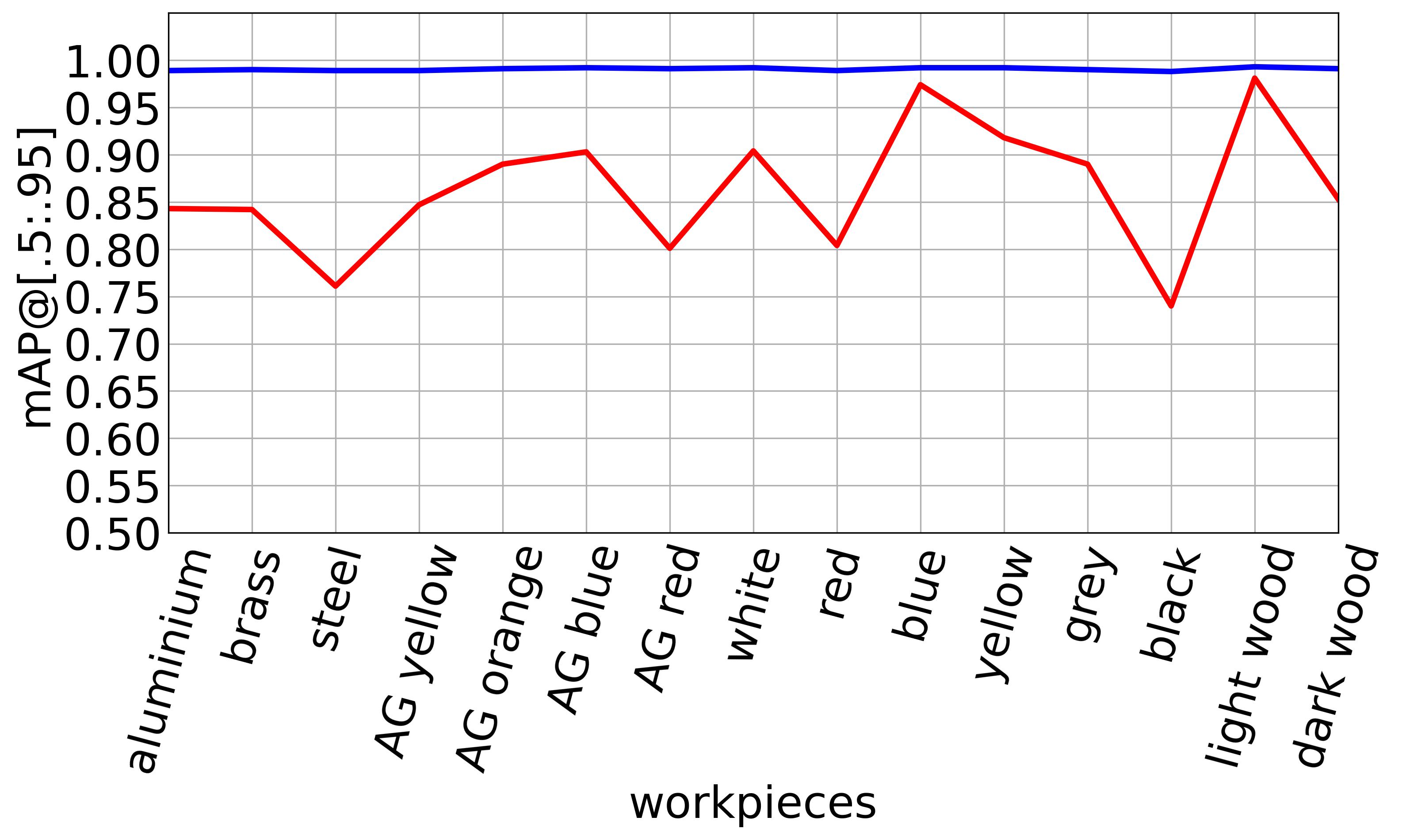}
    \caption{Results for the YOLOv8x multi-class model, trained on all the images at once, (blue) training mAP, (red) evaluation mAP)}
    \label{fig:MulticlassYOLOv8x}
\end{figure}

The YOLOv8 models that have previously been evaluated have only been trained on one class each. In reality, however, it is often necessary for a (real-time) object detection framework to be able to locate and to classify different objects. Therefore, we trained a YOLOv8n and YOLOv8x model on all 15 workpieces at once, including all (72,000 in total) training images. The mAP@[.5:.95] results for training and evaluation are displayed in Fig.\ \ref{fig:MulticlassYOLOv8n} (YOLOv8n) and  Fig.\ \ref{fig:MulticlassYOLOv8x} (YOLOv8x). 

Both models are fairly well trained, with an mAP@[.5:.95] of about 0.95, while the larger model, like in the single-class case, shows better performance. In contrast to where the evaluation of YOLOv8x was worse than YOLOv8n in the single-class case, we now see the trend one may have expected. Comparing both evaluation mAPs, we can observe that the general trend of both models is similar and reminds of the single-class results. The red coloured, the steel and especially the black workpieces show the worst performance. Besides an overall mAP@[.5:.95] increase, the blue and the light wood workpiece show an mAP@[.5:.95] of over 0.95. In total, we cannot find characteristics of over- or underfitting in the multi-class training. Therefore, the trend we can observe in Fig.\ \ref{fig:MulticlassYOLOv8n} Fig.\ \ref{fig:MulticlassYOLOv8x} seems to be reliable.

\section{Summary and conclusion}

We have trained a total of 92 YOLOv8 models, which divide into three different sized datasets and two model variants (nano and extra large) for each of the 15 workpieces and two times YOLOv8n and YOLOv8x on all training images in the multi-class case. For both the training and evaluation, datasets were constructed in a way that the only changing parameter was the appearance/material of the cylindrical workpieces. 

The analysis of the results has shown that the workpieces made of yellow and white plastic, orange acrylic glass and light-coloured wood could be learned very well. The detection of aluminium, steel, black and red plastic, as well as red acrylic glass was problematic. In general, background objects of the learning factory were sometimes recognised and had an impact on the detection. It seems that both colour and contour are not the only factors that are important, the reflecting characteristic is also relevant. We highlighted some of the difficulties in object/background interaction, illuminating also that some questions about, for instance, the possible overfitting are not easily addressed. However, the presented study involves some limitations. Based on the amount of workpieces (15) and the dataset creation approach, only a limited amount of original recorded images is currently available. Although we find our conclusions based on the results to be solid, the findings require additional investigation and validation with, e.g., additional images. 

In future work, we aim to enhance the explainability of the approach with the focus on understanding which features may be more salient for detection in complex environments as in this paper. We will also use the created datasets to investigate and validate, whether the findings in this paper are applicable to other object detection models, such as R-CNN. In addition, characteristics of the datasets will be investigated and highlighted, in order to support the impact of the composition on the findings. The dataset can also be extended with different shaped workpieces. 

\section*{Acknowledgement}
This work was funded by the Bundesministerium für Bildung und Forschung (BMBF) within the project \emph{KI@MINT} ("AI-Lab").
%
%

\end{document}